\documentclass{article}

\usepackage{arxiv}
\usepackage{xskak}
\usepackage{chessfss}

\usepackage[T1]{fontenc}
\usepackage[utf8x]{inputenc}
\usepackage{csquotes}
\usepackage[hidelinks]{hyperref}       
\usepackage{url}            
\usepackage{booktabs}       
\usepackage{amsfonts}       
\usepackage{nicefrac}       
\usepackage{microtype}      
\usepackage{graphicx}
\usepackage{natbib}
\usepackage{doi}

\title{You shall know a piece by the company it keeps. Chess plays as a data for word2vec models}

\author{\href{https://orcid.org/0000-0002-9099-0436}{\includegraphics[scale=0.06]{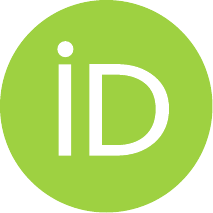}\hspace{1mm}Boris Orekhov}, \\
	School of Linguistics
	HSE University,\\
	Institute of Russian Literature (Pushkin House)\\
	Russian Academy of Sciences \\
	\texttt{borekhov@hse.ru} 
}

\renewcommand{\shorttitle}{You shall know a piece by the company it keeps. Chess plays as a data for word2vec models}

\hypersetup{
pdftitle={You shall know a piece by the company it keeps. Chess plays as a data for word2vec models},
pdfsubject={cs.CL, cs.AI},
pdfauthor={Boris Orekhov},
pdfkeywords={chess, word2vec, superlinguistics, vector space models},
}

\begin{document}
\maketitle

\begin{abstract}
	In this paper, I apply linguistic methods of analysis to non-linguistic data, chess plays, metaphorically equating one with the other and seeking analogies. Chess game notations are also a kind of text, and one can consider the records of moves or positions of pieces as words and statements in a certain language. In this article I show how word embeddings (word2vec) can work on chess game texts instead of natural language texts. I don't see how this representation of chess data can be used productively. It's unlikely that these vector models will help engines or people choose the best move. But in a purely academic sense, it's clear that such methods of information representation capture something important about the very nature of the game, which doesn't necessarily lead to a win.

\end{abstract}


\section{Introduction}

In this paper, I apply linguistic methods of analysis to non-linguistic data, metaphorically equating one with the other and seeking analogies. The productivity of this approach has been proven within the field of \href{https://www.hf.uio.no/iln/english/research/groups/super-linguistics/}{Super Linguistics}.

I argue that developed by computational linguists word embeddings (made with the algorithm word2vec) can shed light on the features of chess moves.

Recently, computational linguistics has made a great progress in natural language processing (NLP). Within this field, tools have been developed for machine analysis of morphology, syntax and semantics. Computational linguistics became a stand-alone research area with its conferences (\href{https://aclanthology.org/}{weblink}) and actual fields. The experts have developed general principles of text analysis, optimal methods of word counting. e.g. term frequency, inverse document frequency (tf-idf) is used instead of simple word frequency, for finding collocations Pointwise Mutual Information is used, etc. Computational linguistics technologies are successfully implemented in the commercial products in the field of information retrieval \citep{roy2018using}, electronic marketing \citep{woltmann2018modeling}, modern user interfaces (voice assistants, chat bots), and so on.

At the same time, there are works that show the possibility of applying NLP tools to non-NL data. The idea of word embeddings proved to be particularly productive outside the text data. Word embedding is a representation of the semantics of a word as a vector in a multidimensional space, the measurements of which are determined by the contexts in which the word appears in the texts. \citep{baroni2014dont, levy2015improving}. The main technological breakthrough of word embeddings was that the number of measurements for each word can be relatively small, much less than the number of words in the dictionary.

Word embeddings are a technical implementation of the linguistic idea of distributional semantics. Distributional semantics is the concept that the meaning of a word is revealed through its context, and by examining the surrounding words, we can infer its semantics. For example, we can assume that in the sentence "he returned home after three weeks" and in the sentence "he returned home after three years," the words "week" and "year" are similar in meaning because their verbal surroundings are the same. This is indeed the case; both words denote a length of time, though not of equal duration.

In computational linguistics, vectorization algorithms (the most popular being word2vec) have been developed, which allow us to represent each word (in our case, "week" and "year") as a vector in n-dimensional space. These dimensions are precisely the neighboring words in whose context the word of interest appears in real texts. The vector coordinates are the frequency with which these words appear in the same sentence as our word (i.e., how often the words "he," "return," "home," "three" occur in the same context with "year" and "week").

As a result, we obtain a vector model for the words in the corpus we are interested in, and this model can be used for various interesting manipulations that can be performed with vectors in mathematics. For instance, we can find the vectors closest to a given one. If we are interested in the word "year," we calculate the vectors closest to it and find that these vectors correspond to the words "week," "month," "half-year," etc. "Closest" means that the cosine distance between these vectors is minimal. These models are useful for solving research problems \citep{orekhov2023individual}.
For NLP word embeddings were extremely productive, they allowed to describe the semantics of a word using machine methods, based only on a rather large amount of text. But it is important that the ideas formulated in this area were extended to the structures of graphs \citep{narayanan2017graph2vec}, practical medicine data \citep{stella2015word2vec}, image processing and pattern recognition \cite{li2017deep} and finally for categorical data in statistics in general \citep{wen2016cat2vec}. One of the most impressive achievements of embeddings outside the word area was the automatic reconstruction of the periodic table of elements. \citep{than2018ai}
Another curious spread of achievements of computational linguistics and artificial intelligence beyond the language area was the experiment with GPT-2 neural network. This neural network architecture was represented by OpenAI. GPT-2 became known due to the fact that having trained on a large collection of texts, it is able to generate seemingly meaningful new texts that are almost indistinguishable from human ones. In the experiment I talk about, \citep{alexander2020very} GPT-2 was trained not on texts in natural language, but on texts of chess games. As a result, such a training sample was enough for GPT-2 to make reasonable moves and have a game with a human chess player.
In this article I will show how word embeddings can work on chess game texts instead of natural language texts. I found no publications on these topics (I only saw one semi-study programming exercise on a similar topic: \href{https://github.com/queirozfcom/chess-word2vec}{weblink}).

Chess moves, like words in natural language, receive their interpretation from context. My main hypothesis was the following: chess moves can be classified by their neighbors in the game into different groups.

\section{Data}

Chess game notations are also a kind of text, and one can consider the records of moves or positions of pieces as words and statements in a certain language.
According to the rules of the PGN (portable game notation) format, an uppercase letter denotes a white piece, while a lowercase letter denotes a black piece. The letters themselves are obvious: R - rook \rook, N - knight \knight, B - bishop \bishop, etc. Castling is denoted by the move of the king (the movement of the rook is implied by default).

An example of a chess game recorded in PGN format looks like this:

\begin{verbatim}
Pe2e4 pc7c5 Ng1f3 pd7d6 Pd2d4 pc5d4 Nf3d4 ng8f6 Nb1c3 pa7a6 Bf1e2 pe7e6 Ke1g1 bf8e7 Pf2f4
ke8g8 Bc1e3 nb8c6 Kg1h1 bc8d7 Pa2a4 ra8c8 Nd4b3 nc6a5 Nb3d2 bd7c6 Be2d3 pd6d5 Pe4e5 pd5d4
Be3d4 qd8d4 Pe5f6 be7b4 Pf6g7 qd4g7 Nd2f3 bb4c3 Pb2c3 bc6d5 Qd1e2 rc8c3 Ra1e1 na5c6 Qe2d2
kg8h8 Re1e3 nc6b4 Nf3e5 nb4d3 Pc2d3 rf8c8 Re3g3 rc3c2 Qd2b4 rc2g2 Ne5f7 kh8g8 Nf7h6 kg8h8
Nh6f7 kh8g8 Nf7h6
\end{verbatim}

etc.

Here, the first letter in each "word" denotes the piece and its color, the next two symbols indicate the square from which the piece moves, and the last two symbols indicate the square to which the piece moves. Thus, \texttt{Pe2e4} is a move by a white pawn (uppercase letter, so the piece is white, the letter P indicates a pawn) from square \texttt{e2} to square \texttt{e4}. For more details see \citep{boutell2003portable}. 

One could build vector models based on game texts that reflect the moves, as in the example above, but this is somewhat controversial because, with this approach, the "context" is lost, meaning the position in which the move is made. Considering moves separately from the position is a strange idea for chess. Moves and their function in the game do not exist independently of the position. However, this is similarly strange when it comes to words, when we look at them out of context. 

Words have no meaning outside of context and acquire it only as part of a certain statement. We do not always realize this because we are used to seeing words on dictionary pages. Dictionaries try to reflect different meanings but can never do so completely. When we think of a polysemous word, we only recall one, the most obvious meaning, while we temporarily forget the others. Similarly, chess moves in vector models will have to be handled.

Another way to linguistically reconceptualize the moves of a chess game can lie within the framework of the concept of topic and comment. The topic, or theme, of a sentence is what is being talked about, and the comment (rheme or focus) is what is being said about the topic.

In our "text," the words express only the rheme, while the theme is lost. Or, we can look at the problem differently: each word simultaneously expresses both the theme (the name and color of the piece, the square it stood on) and the rheme (the square to which the piece is moved as a result of the move). Thus, differences in the structure of linguistic and chess material become apparent.

At the same time, building vector models based on the list of moves in a game does not seem entirely absurd, because with this approach, we get moves translated into vectors that "belong to the same company," meaning they regularly occur in the same game situations.

Modern methods of natural language processing and information retrieval depend heavily on large collections of text. I used a collection of records of 5,400,137 chess games (approx 840 mln moves, \href{https://github.com/rozim/ChessData}{weblink}). These are mainly games played at a high international level.

Based on this data, I trained two types of models.

Type 1 is a model "based on moves." That is, a listing of a game such as \texttt{Pe2e4} \texttt{pc7c5} \texttt{Ng1f3} \texttt{pd7d6} \texttt{Bf1b5} \texttt{bc8d7} \texttt{Bb5d7} \texttt{qd8d7} and so on is taken and considered as a "sentence," with each move in it being a separate word.

Type 2 is a model "based on positions." Here, each move is considered as a "sentence," where the words will be both the move itself (\texttt{->Bc1h6}) and all the squares occupied by pieces (\texttt{Pa2} \texttt{Pb2} \texttt{Pc3} and so on). The position on the board is the context for the move, just as words form the context for a word in a statement in a natural language. Therefore, I prepared the source material differently for the models of the second type. Usually, when building a vector model on a language corpus, this corpus is divided into sentences, and the vectorization captures only the context of one sentence. For chess statements, I broke down the game notations into individual moves, and I got "sentences" of this kind:

\begin{enumerate}
\item \texttt{ra8} \texttt{nb8} \texttt{bc8} \texttt{qd8} \texttt{ke8} \texttt{bf8} \texttt{ng8} \texttt{rh8} \texttt{pa7} \texttt{pb7} \texttt{pc7} \texttt{pd7} \texttt{pe7} \texttt{pf7} \texttt{pg7} \texttt{ph7} \texttt{Pa2} \texttt{Pb2} \texttt{Pc2} \texttt{Pd2} \texttt{Pe2} \texttt{Pf2} \texttt{Pg2} \texttt{Ph2} \texttt{Ra1} \texttt{Nb1} \texttt{Bc1} \texttt{Qd1} \texttt{Ke1} \texttt{Bf1} \texttt{Ng1} \texttt{Rh1} \texttt{->Pe2e4}
\item \texttt{ra8} \texttt{nb8} \texttt{bc8} \texttt{qd8} \texttt{ke8} \texttt{bf8} \texttt{ng8} \texttt{rh8} \texttt{pa7} \texttt{pb7} \texttt{pc7} \texttt{pd7} \texttt{pe7} \texttt{pf7} \texttt{pg7} \texttt{ph7} \texttt{Pe4} \texttt{Pa2} \texttt{Pb2} \texttt{Pc2} \texttt{Pd2} \texttt{Pf2} \texttt{Pg2} \texttt{Ph2} \texttt{Ra1} \texttt{Nb1} \texttt{Bc1} \texttt{Qd1} \texttt{Ke1} \texttt{Bf1} \texttt{Ng1} \texttt{Rh1} \texttt{->pc7c5}
\item \texttt{ra8} \texttt{nb8} \texttt{bc8} \texttt{qd8} \texttt{ke8} \texttt{bf8} \texttt{ng8} \texttt{rh8} \texttt{pa7} \texttt{pb7} \texttt{pd7} \texttt{pe7} \texttt{pf7} \texttt{pg7} \texttt{ph7} \texttt{pc5} \texttt{Pe4} \texttt{Pa2} \texttt{Pb2} \texttt{Pc2} \texttt{Pd2} \texttt{Pf2} \texttt{Pg2} \texttt{Ph2} \texttt{Ra1} \texttt{Nb1} \texttt{Bc1} \texttt{Qd1} \texttt{Ke1} \texttt{Bf1} \texttt{Ng1} \texttt{Rh1} \texttt{->Ng1f3}
\end{enumerate}

etc.

\newchessgame
\mainline{1.e4 c5 2.Nf3}
\chessboard

Here, the position of the pieces on the board is recorded first (\texttt{ra8}, \texttt{nb8}, \texttt{bc8}...), followed by the move made by the player: \texttt{->Pe2e4.} To avoid confusing the "words" that denote the position of a piece on the board with the "words" that denote a move, I added a prefix in the form of the symbol \texttt{->} to each move.

Thus, both the positions of the pieces on the board and the move are all "words" of one "sentence." 

The algorithm can be given a window of observation, meaning how many words to the left and right of the current one we should consider. To account for the entire board, I set the window to 32, so when calculating the vector of a move, all positions of the pieces on the board will be taken into account.

List of models for the study:

\begin{enumerate}
\item \texttt{moves\_texts.model}, Type 1 model, all moves of all games in the collection (move $=$ "word," game = "sentence").
\item \texttt{lemmatized\_moves\_texts.model}, Type 1 model, all moves of all games (move $=$ "word," game $=$ "sentence"), but the square from which the piece moves is excluded from the move. This is a variant of "stemming."
\item \texttt{white\_moves.model}, Type 2 model, includes only white's moves along with positions (move and all pieces on the board $=$ "words," position $+$ move $=$ "sentence").
\item \texttt{black\_moves.model}, Type 2 model, same as above but for black.
\item \texttt{debut\_moves.model}, Type 1 model, the "sentence" is the truncated segment of the beginning of the game up to the 12th move. It should reflect only the opening moves of the game.
\item \texttt{debut\_positions.model}, Type 2 model, only moves up to the 12th in the game along with their positions. Only white's moves are included; black's moves are excluded.
\item \texttt{mittel\_moves.model}, Type 1 model, only moves from the 13th to the 30th. Presumably the middlegame.
\item \texttt{mittel\_positions.model}, Type 2 model, moves along with positions from the 13th to the 30th. Only white.
\item \texttt{endgame\_moves.model}, Type 1 model, moves from the 31st to the end of the game.
\item \texttt{endgame\_positions.model}, Type 2 model for moves and positions, starting from the 31st in the game. Only white.
\item \texttt{moves\_pos.model}, Type 1 model, differs from model \texttt{moves\_texts.model} only in that each move is added with a "part-of-speech tag." There are only two tags: \texttt{\_CAP} and \texttt{\_N}. Thus, for each move, there can be two entries: \texttt{Bc1h6\_N} and \texttt{Bc1h6\_CAP}. \texttt{Bc1h6\_CAP} means the bishop moves to square \texttt{h6} with a capture, while \texttt{Bc1h6\_N} is the same move without a capture. Thus, the linguistic idea of parts of speech is transferred to the text of chess games: there is a part of speech for a move with a capture and a part of speech for a move without a capture.
\item \texttt{positions\_pos.model}, Type 2 model with part-of-speech tags. So it can be \texttt{->Bc1h6\_N} and \texttt{->Bc1h6\_CAP}. Only white.
\item \texttt{queens\_moves.model}, Type 1 model, moves in the game until there is at least one queen on the board. I did not exclude cases when a queen reappears on the board in the endgame after a pawn promotion.
\item \texttt{no\_queens\_moves.model}, Type 1 model, moves in the game from the moment both queens disappear from the board.
\item \texttt{queens\_positions.model}, Type 2 model, moves $+$ their positions when there is at least one queen on the board. Only white.
\item \texttt{no\_queens\_positions.model}, Type 2 model, moves $+$ their positions when both queens have disappeared from the board. Only white.
\item \texttt{positions\_moves\_pro.model}, A model combining types 1 and 2. This model includes the position of the move, the move itself, and three moves before and after it. Perhaps such a model will better reflect the player's strategy (aggressive or passive play, attacking or positional). Only white.
\item \texttt{result\_moves.model}, Type 1 model, includes only decisive games.
\item \texttt{tied\_moves.model}, Type 1 model, includes only drawn games.
\end{enumerate}

All the models are published on Huggingface \citep{orekhov2024w2v}.

\section{Results}

First, we need to ensure that the models truly reflect chess reality and do not contain random sequences of moves from a chess perspective. Vector models allow us to find similar words (in our case, moves). Let's try working with the basic model \texttt{moves\_texts.model}. The model contains $7749$ words (moves). Here are a few random examples. First, the move is listed (highlighted in bold) for which we are searching for the closest move vectors, followed by the "synonym" moves (usually referred to as quasi-synonyms), and the cosine distance between the vectors is indicated (ranging from 0 to 1, the higher the number, the closer the vectors are to each other):

\textbf{Pe2e4}, \pawn e2e4

\begin{verbatim}
Pe2e3 0.4319874346256256
Ng1f3 0.3651273250579834
Pd2d4 0.36322540044784546
Pe3e4 0.34142181277275085
ng8f6 0.3036213517189026
pd7d5 0.29061219096183777
pc7c5 0.28487297892570496
Pc2c4 0.2729674279689789
Pc3d4 0.2568834125995636
ke8g8 0.24415579438209534
\end{verbatim}

\textbf{Bf1b5}, \bishop f1b5

\begin{verbatim}
Bf1d3 0.5939557552337646
Bf1g2 0.5874636769294739
Bf1e2 0.5806597471237183
Bf1c4 0.5764033794403076
Bc4b5 0.518180251121521
Bd3b5 0.4963855743408203
Be2b5 0.4832766652107239
Ba4b5 0.36604568362236023
Bf1a6 0.3487110435962677
Bf1h3 0.3368876576423645
\end{verbatim}

\textbf{Rf8f4}, \rook f8f4

\begin{verbatim}
Rf8b8 0.6256532669067383
Rf8d8 0.6229713559150696
Rf8a8 0.6161673069000244
Rf8h8 0.6064906716346741
Rf8c8 0.6060029864311218
Rf8e8 0.5938045382499695
Rf8g8 0.5920660495758057
Rf8f3 0.5568860173225403
Rf8f6 0.5522985458374023
Rf8f2 0.5472667813301086
\end{verbatim}

From these lists, it is clear that the vectors in the model are not distributed randomly. The moves closest to a given move are usually those made by the same piece. This is not obvious in chess, as different continuation options can exist in similar game situations (positions are not considered in this model). However, linguistically, this is expected because quasi-synonyms are usually words of the same part of speech or even cognates.

Furthermore, to find the moves closest in "semantics" to a given move, we can calculate the cosine distance between any two arbitrary moves in the model. For example, for \texttt{Pe2e4} and \texttt{pe7e5}, the distance is $0.1094483$. This is a minimal value, resulting because the context for the opening move \texttt{Pe2e4} includes all possible continuations, whereas \texttt{pe7e5} is relatively local compared to this variety.

Another available function is "find the odd one out." In a linguistic vector model, if we propose a series of words like "apple," "pear," "grape," "banana," "orange," and "cucumber," the model will indicate that "cucumber" is the odd one out. This is because "cucumber" is typically used in a different context than the other words, so its vector has noticeably different coordinates.

Let's try the same with moves. We'll take some common moves for traditional openings, add a move that is typically played not in the opening but in the endgame, and see if the model can identify the odd one out. The series is:

\texttt{Pe2e4, Ng1f3, Pd2d4, Bf1c4, Nb1c3, Pb4b5}

Here, all the moves are from openings (the Four Knights and the Bishop's Opening), except for the pawn moving to the edge of the board on the b-file. Yes, the model immediately highlights this move and indicates that it is the odd one out in this series. 

Thus, the results are somewhat meaningful. On the other hand, the move Bf1c4 can be made outside the opening. Among the remaining moves, this one is the least similar to an opening move. If we remove the move Pb4b5, what will be the odd one out? The model indicates that among the series 

\pawn e2e4, \knight g1f3, \pawn d2d4, \bishop f1c4, \knight b1c3

the odd one out is \bishop f1c4. This also suggests the results are meaningful: the model distinguishes moves made in different game situations.

Another function is solving proportions. The most famous example for linguistic vector models is to take the word "king," add the vector for "woman," subtract the vector for "man," and you get "queen."

    Let's try to do this with moves. We'll take typical opening moves for white and add a move from the same opening for black. King's Gambit: \texttt{Pe2e4} $-$ \texttt{Pf2f4} $+$ \texttt{pe7e5}. The model suggests the move \texttt{pe7e6}. This is also an opening move that appears, for example, in the French Defense: \texttt{Pe2e4, pe7e6, Pd2d4, pd7d5}…

\newchessgame
\mainline{1.e4 e6 2.d4 d5}
\chessboard

Now let's work with the model that includes not only moves but the position on the board (Type 2 model): \texttt{positions\_moves\_pro.model}. It includes $4946$ words. The closest quasi-synonym neighbors are:

\textbf{->Pe2e4}, \pawn e2e4

\begin{verbatim}
->Pe2e3 0.4605116844177246
->Pe3e4 0.3592767119407654
->Bf1g2 0.27666473388671875
->Pe2f3 0.24844536185264587
->Pd2d4 0.2099190503358841
->Pc2c4 0.19693881273269653
->Ng1f3 0.1944185346364975
->Pg2g3 0.1831442415714264
->Ke1g1 0.17580436170101166
be2 0.17578239738941193
\end{verbatim}

\textbf{->Bf1b5}, \bishop f1b5

\begin{verbatim}
->Bf1c4 0.7282129526138306
->Bf1d3 0.6547436714172363
->Bf1e2 0.6298115253448486
->Bf1a6 0.5582098364830017
->Bf1h3 0.3876230716705322
->Be2b5 0.36358100175857544
->Bd3b5 0.30369845032691956
->Bc4b5 0.3022392988204956
->Bf1g2 0.29600590467453003
->Pd2d4 0.232324481010437
\end{verbatim}

The closest move to \pawn \texttt{e2e4} turned out to be \pawn \texttt{e2e3}, which is logical: this is also a way to start the game, and in both cases, the pieces are in similar positions.

Since in this model "words" are not only moves but also piece positions, we can do the same with positions as we do with moves. For example, we can find quasi-synonyms for the position of a white pawn on \texttt{e2}:

\begin{verbatim}
Pe3 0.5342991948127747
Pe4 0.47625163197517395
Pe5 0.3307723104953766
->Pe3d4 0.26243242621421814
->Pe4d5 0.2563534677028656
->Re1e5 0.22984528541564941
->Re1e8 0.22572621703147888
->Re1e6 0.20842188596725464
->Bf1c4 0.2069925218820572
->Re1e7 0.20322373509407043
\end{verbatim}

The closest neighbors of the vector for the white pawn in its initial position on \texttt{e2} are the vectors for the white pawn after the first or second move—on \texttt{e3} or \texttt{e4}. Indeed, when the pawn is on these squares, the other pieces on the board are positioned similarly, representing a very similar "context" for the "statement."

So far, we have looked at individual moves in the models and tried to understand something from their nearest neighbors. Now, I propose we look at the models as a whole to see how all the moves are positioned relative to each other.

I created a visualization of all the moves in all the models. For visualization, I used the tSNE algorithm, a dimensionality reduction algorithm. The idea is that when we have a model with all the moves, these moves represent vectors in a multi-dimensional space. Our models have between 300 and 500 dimensions, making it impossible to visualize. tSNE takes all the dimensions present in the model and compresses them into two, projecting the entire multi-dimensional space onto a plane, representing it as if all our moves existed in a two-dimensional space. 

In this process, move-vectors turn into points on the plane. Those with similar dimensions are placed close to each other in the visualization. If they have similar dimensions, it means they are very likely to be neighbors (quasi-synonyms). Thus, in the visualization, synonym moves cluster into "bundles," areas where points are densely packed together. Consequently, vectors with different dimensions end up in different clusters. The relationships between these clusters are then determined by the settings of the algorithm.

I created diagrams for all our models with different perplexity values (5, 30, 50), which is the main setting for the tSNE algorithm. Higher perplexity causes clusters to become more distinct from each other. More details can be found \href{https://distill.pub/2016/misread-tsne/}{here}. 

Not all points on the graphs are labeled (there are several thousand, labeling each one would be too cluttered), but every third point is labeled. Additionally, I made color versions of the graphs where each piece's move is marked with its own color. This shows that moves with the same piece tend to cluster together. The labels show that within one cluster, moves with the same piece from the same square are grouped. This is similar to what we see when looking at quasi-synonyms of moves. This is already a surprising result because while in type 2 models (built on positions), it is clear from which square a piece moves (other moves are impossible), in type 1 models (built only on moves), this is not so obvious. How does the model know that moves with the same piece in similar game contexts will be similar?

\begin{figure}
	\centering
	\includegraphics[width=0.9\textwidth]{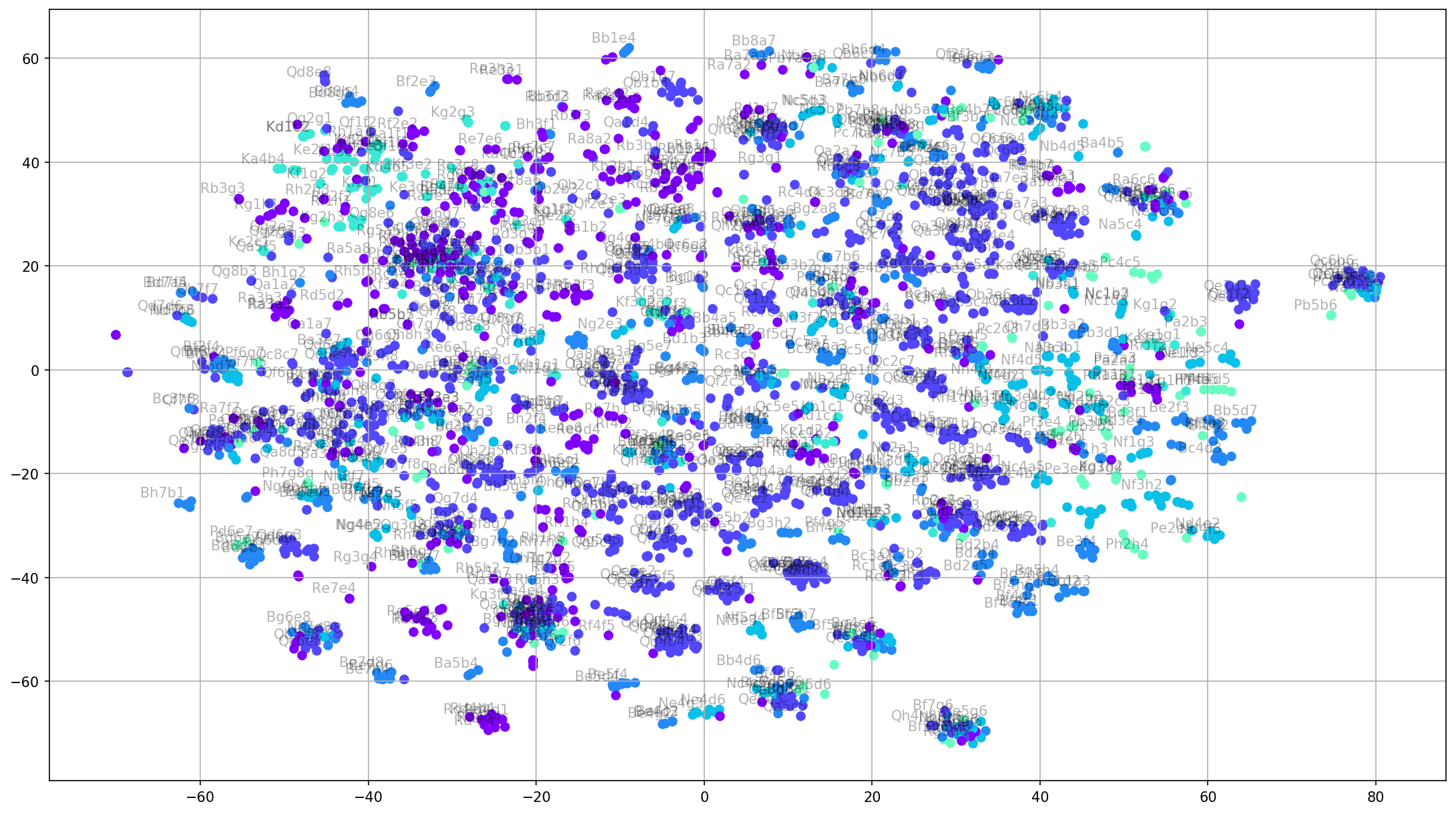}
	\caption{tSNE visualisation of the moves from the debut\_moves\_white.model with perplexity 5}
	\label{fig:fig1}
\end{figure}

\begin{figure}
	\centering
	\includegraphics[width=0.9\textwidth]{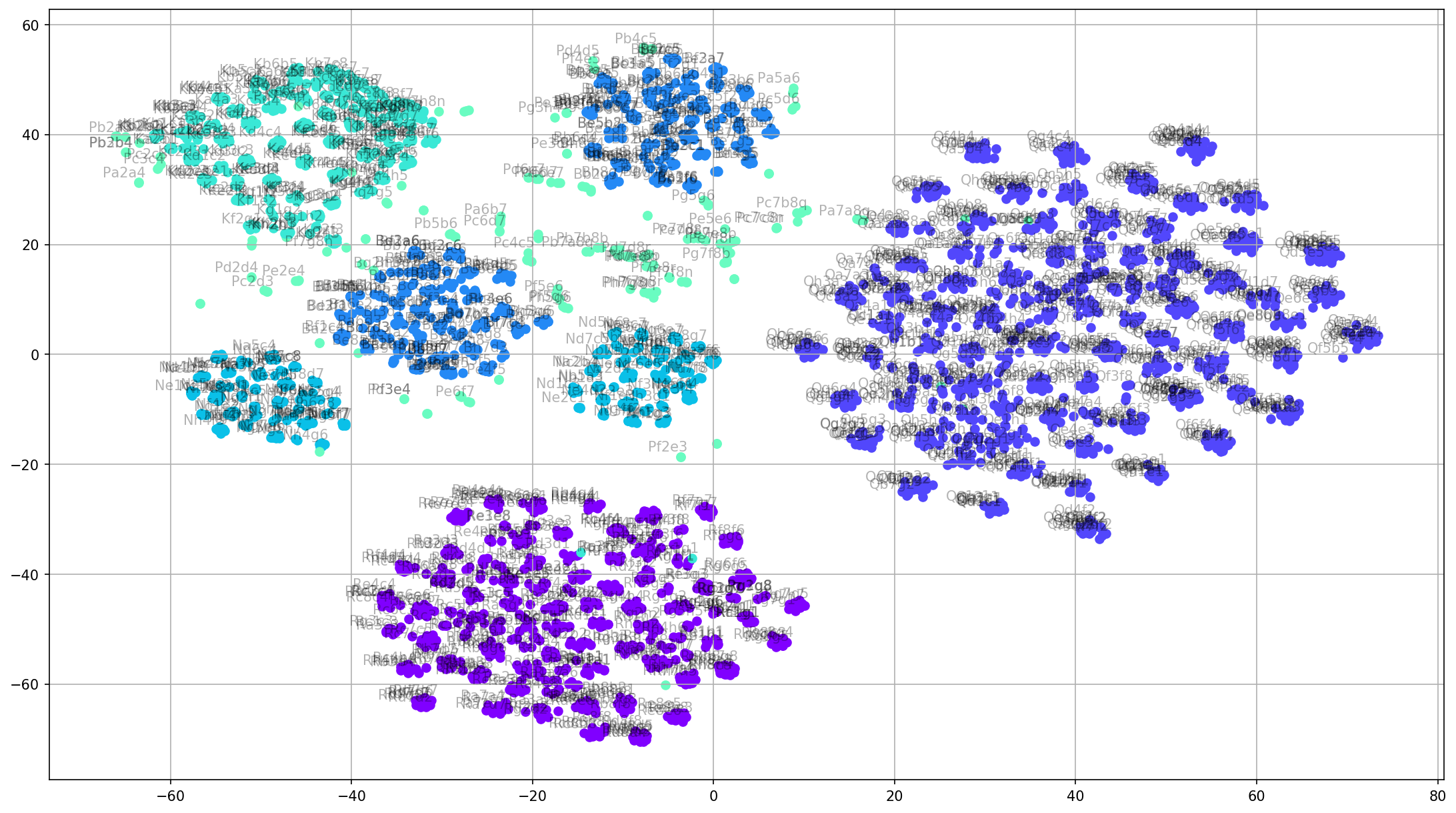}
	\caption{tSNE visualisation of the moves from the endgame\_white.model with perplexity 30}
	\label{fig:fig2}
\end{figure}

The analysis of the visual representation yields the following result: opening moves do not cluster well (do not form distinct clusters): fig. \ref{fig:fig1} This seems as it should be: all openings are similar, the same moves are made, so there are no pronounced trends. However, endgame moves cluster very clearly, as in a textbook—distinctly by each piece: fig. \ref{fig:fig2} Chess players might be interested in exceptions to this principle.

\begin{figure}
	\centering
	\includegraphics[width=0.9\textwidth]{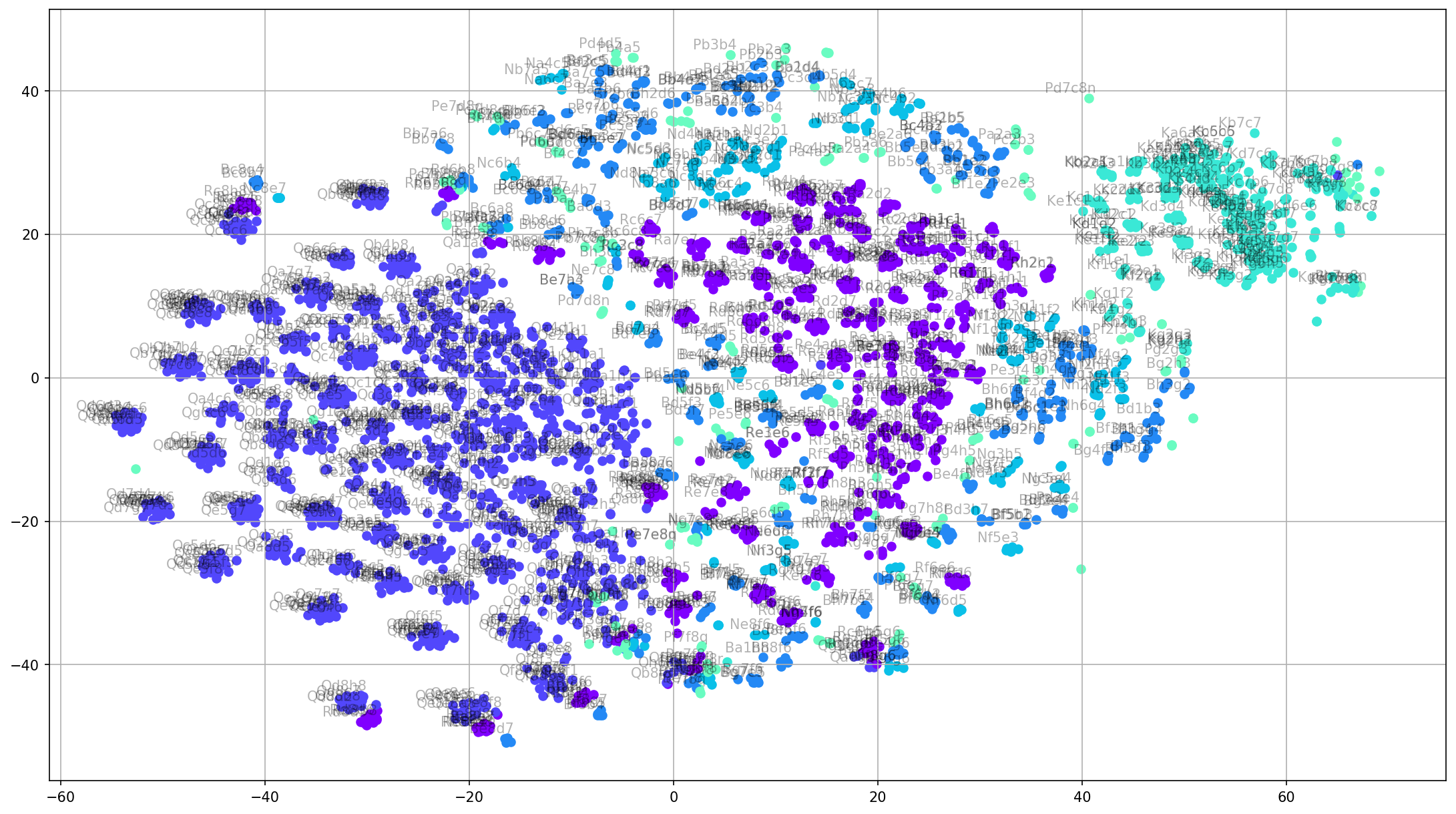}
	\caption{tSNE visualisation of the moves from the mittel\_white.model with perplexity 50}
	\label{fig:fig3}
\end{figure}

Midgame moves fall somewhere in between—there are clear clusters, but they are closer together than the clusters of endgame moves. The king and queen "live" separately, while the center of the graph is occupied by the rook. Moves with the rook are equidistant from moves with all other pieces: fig. \ref{fig:fig3}.

\begin{figure}
	\centering
	\includegraphics[width=0.9\textwidth]{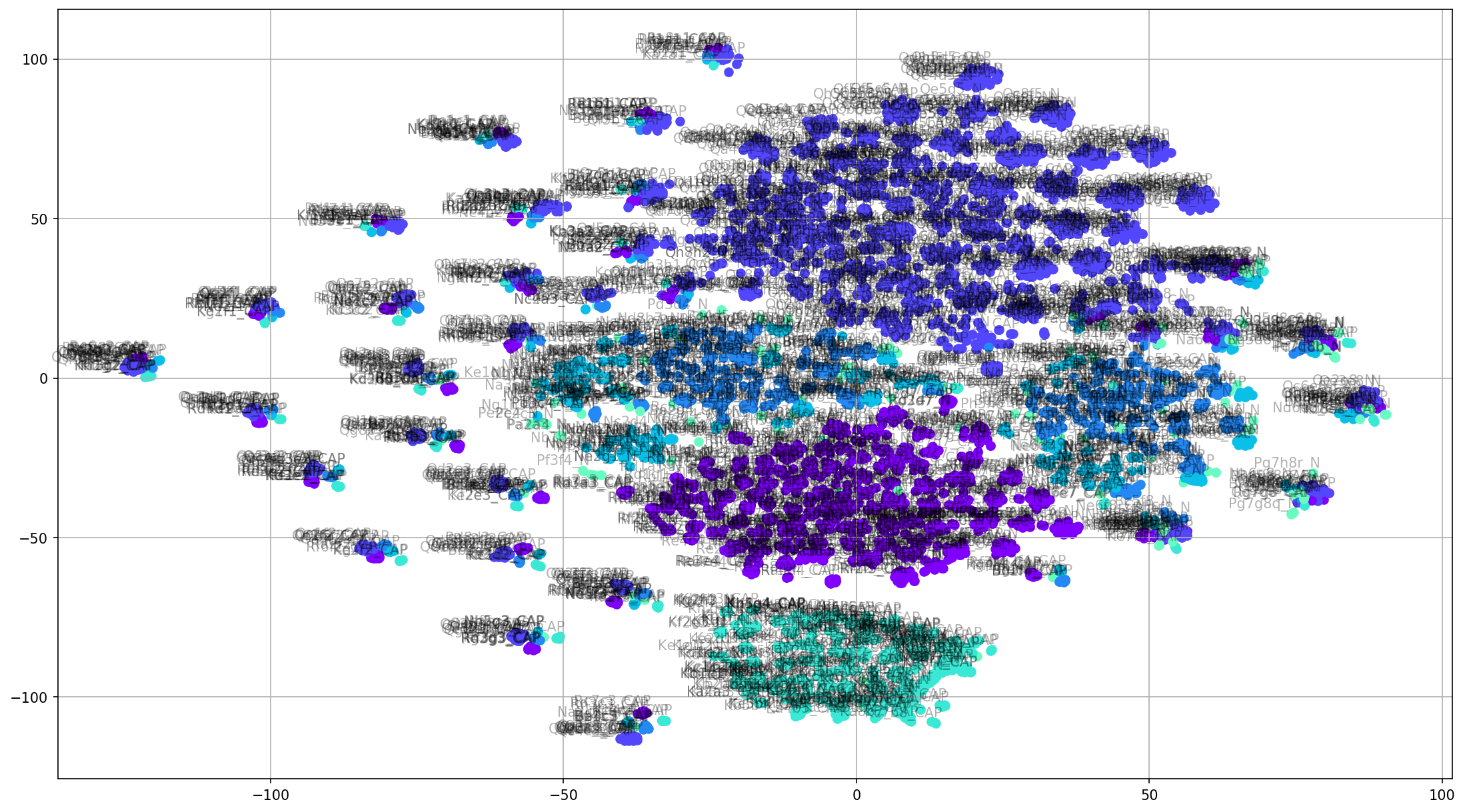}
	\caption{tSNE visualisation of the moves from the moves\_pos\_white.model with perplexity 30}
	\label{fig:fig4}
\end{figure}

An interesting result emerged from the model that differentiates between moves with captures (postfix \texttt{\_CAP}) and without captures (postfix \texttt{\_N}): fig. \ref{fig:fig4}. In this model, clusters form based on moves to the same square rather than from the same square. So, whereas in other models moves like \rook \texttt{a1a3} and \rook \texttt{a1a5} would cluster together, here moves like \texttt{kf2e2\_CAP} and \texttt{kf3e2\_CAP} cluster together. Additionally, it is interesting that in this model, there are clusters for white pieces that include different pieces, which is rare for other models.

\begin{figure}
	\centering
	\includegraphics[width=0.9\textwidth]{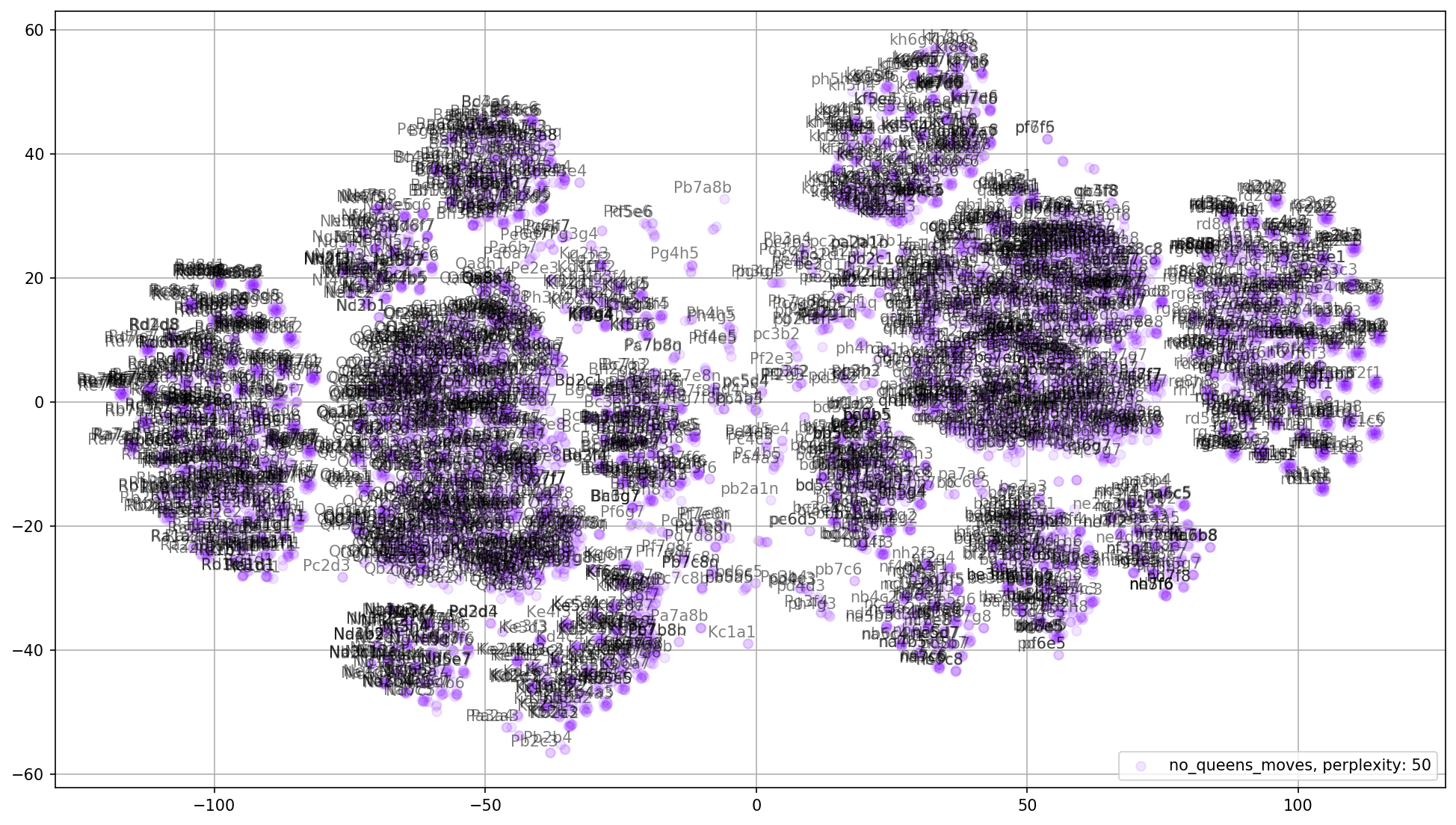}
	\caption{tSNE visualisation of the moves from the no\_queens\_moves.model with perplexity 50}
	\label{fig:fig5}
\end{figure}

A very intriguing fig. \ref{fig:fig5} came from the model reflecting games without queens on the board. This model includes both black and white moves, and they are displayed absolutely symmetrically on the graph, as if in a mirror image.

\begin{figure}
	\centering
	\includegraphics[width=0.9\textwidth]{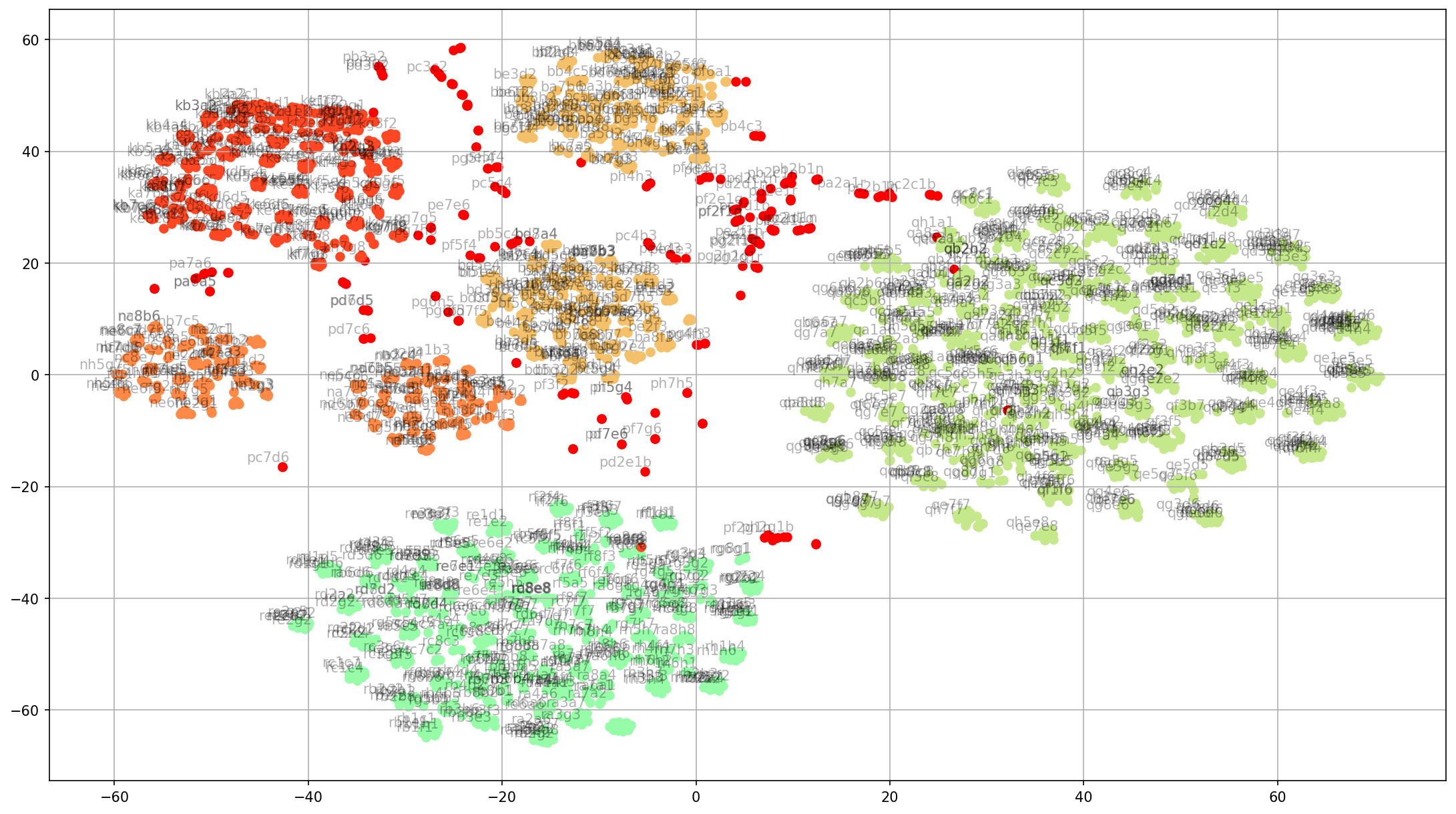}
	\caption{tSNE visualisation of the moves from the endgame\_moves\_black.model with perplexity 30}
	\label{fig:fig6}
\end{figure}

In fig. \ref{fig:fig6}, the white-squared bishop is in a different cluster from the black-squared bishop, which makes sense from a game-play perspective.

\begin{figure}
	\centering
	\includegraphics[width=0.9\textwidth]{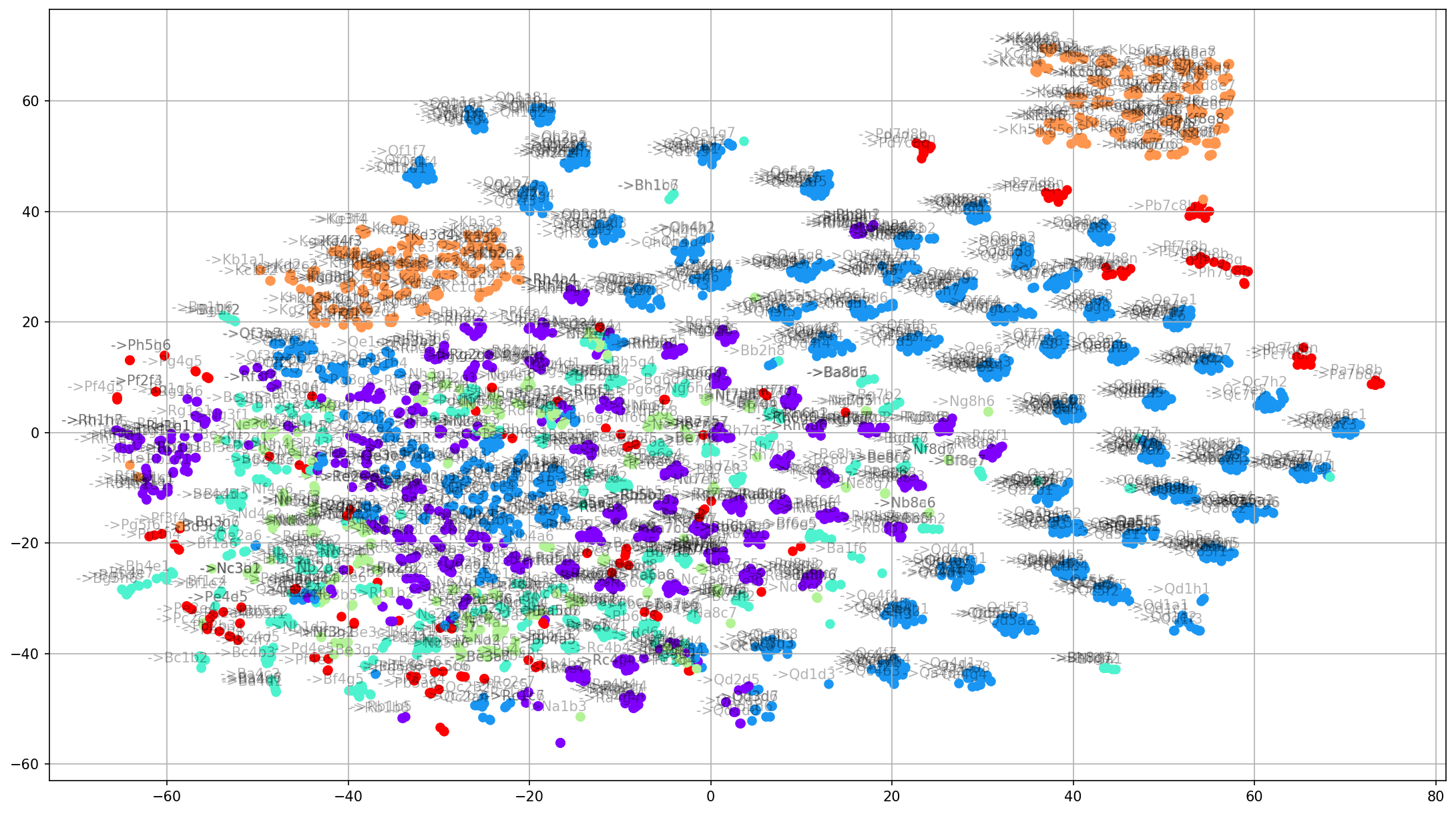}
	\caption{tSNE visualisation of the moves from the positions\_white.model with perplexity 50}
	\label{fig:fig7}
\end{figure}

In fig. \ref{fig:fig7}, there are two clusters for king moves: one cluster for king moves from the late endgame (like \king \texttt{b3c3}) and another cluster for king moves from the middle game or early endgame (like \king \texttt{d8e7}).

We can analyze distant king moves for white in the \texttt{positions\_moves\_pro.model}, which are possible only in the final part of the game:

\textbf{->Kc6b7}, \king c6b7

\begin{verbatim}
->Kc6c7 0.7059599161148071
->Kc6b6 0.6991754770278931
->Kc6b5 0.6284736394882202
->Kc6d5 0.6259784698486328
->Kc6d7 0.6244097352027893
->Ka6b7 0.6211146116256714
->Kc7b7 0.5939313769340515
->Ka7b7 0.5806235671043396
->Kc6d6 0.5764902830123901
->Kb6b7 0.5578444600105286
\end{verbatim}

\textbf{->Kh6g7}, \king h6g7

\begin{verbatim}
->Kh6h7 0.744156002998352
->Kh7g7 0.6991307735443115
->Kg6g7 0.6785054206848145
->Kh6h5 0.6372232437133789
->Kf6g7 0.5794045925140381
->Kh6g5 0.5779032707214355
->Kh7g8 0.566023588180542
->Kg7f7 0.5618337392807007
->Kg6h7 0.5583189129829407
Kh6 0.5513161420822144
\end{verbatim}

Among the quasi-synonyms, there are only similar endgame moves by the white king. This indicates that the model, at the very least, distinguishes between different stages of the game.

Usually, the quasi-synonyms of a move will be moves made by the same piece. But what about the quasi-synonyms of castling? After all, castling can only have two variants. Here is what we found (\texttt{positions\_moves\_pro.model}):

Short castling for white:

\textbf{->Ke1g1}

\begin{verbatim}
->Ke1f1 0.2816329002380371
->Ke1e2 0.268799751996994
->Ke1d1 0.26333796977996826
->Qd1c2 0.2582662105560303
->Ke1f2 0.24763570725917816
->Rh1f1 0.23578758537769318
->Nb1c3 0.2351459413766861
->Qd1b3 0.23325291275978088
->Ke1c1 0.2307348996400833
->Ke1d2 0.2285858690738678
\end{verbatim}

The closest move turns out to be moving the king from the same (initial) square in the same direction: \king \texttt{e1f1}. The other two are moves by the king from the same square to the available squares around it. Among the quasi-synonyms, there is also a rook move that mirrors its movement during castling: \rook \texttt{h1f1}.

Long castling for white:

\textbf{->Ke1c1}

\begin{verbatim}
->Ke1d2 0.4431324601173401
->Ke1e2 0.40111061930656433
->Rh1g1 0.3611656725406647
->Ke1f2 0.35532787442207336
->Ke1d1 0.3478289246559143
->Ke1f1 0.34341734647750854
->Kc1b1 0.3397798240184784
->Ra1d1 0.32468652725219727
->Rh1d1 0.2370266616344452
->Bf1d3 0.23276549577713013
\end{verbatim}

The situation completely repeats itself: the closest moves are the king and rook moves. Thus, the model, based on the available data, has managed to learn that castling involves the movement of both the king and the rook. It is worth noting that for short castling for white, one of the closest quasi-synonyms is long castling, but for long castling, there are no quasi-synonyms corresponding to short castling.

Quasi-synonyms often turn out to be moves where the piece aims to reach the same square as the square of the initial move. For example, if we ask the model for quasi-synonyms for the move Ng3\textbf{f5}, the model will provide us with moves Ne3\textbf{f5}, Nh4\textbf{f5}, Nh6\textbf{f5}, Nd6\textbf{f5}, Nd4\textbf{f5}, meaning all these moves lead to the square f5.

I checked all the moves in the model \texttt{moves\_texts.model} and counted how many of them have at least 3 out of the 10 closest quasi-synonyms as moves of the same piece to the same square. 

It turned out that out of $7749$ moves in the model, $5327$ (68.74\%) meet this condition with at least 3, $4314$ (55.67\%) with at least 4, $3499$ (45.15\%) with at least 5, $2684$ (34.6\%) with at least 6, and $1940$ (25\%) with at least 7.

This is quite a lot. Apparently, one of the key components of "semantics" in the models constructed based on my chosen principles is the tendency of a piece to occupy a specific square.

Here is data on the most similar and least similar moves in the \texttt{moves\_texts.model} (which is built solely on moves, without considering positions):

Closest:

\begin{verbatim}
['Pe3f4', 'Pg3f4'] 0.80376875
['ph7h5', 'ph6h5'] 0.80034935
['Ph2h4', 'Ph3h4'] 0.7996822
['Bg3f4', 'Bh2f4'] 0.7967998
['ka8b7', 'kb8b7'] 0.79391354
['Kb1b2', 'Ka1b2'] 0.7900427
['Kh8g8', 'Kg8h8'] 0.7898132
['Kc1c2', 'Kb1c2'] 0.78719866
['Bh2e5', 'Bg3e5'] 0.7856372
['Kh1g2', 'Kg1g2'] 0.78496265
['kg1h1', 'kh1g1'] 0.7847194
['Ba2d5', 'Bb3d5'] 0.78321075
['Ka8b8', 'Kb8a8'] 0.7824604
['kg8g7', 'kh8g7'] 0.78082377
['Be5g3', 'Be5h2'] 0.78032154
['Ph3g4', 'Pf3g4'] 0.779487
['ka1b1', 'ka2b1'] 0.776601
['Pb3c4', 'Pd3c4'] 0.7763526
['Kb1a2', 'Ka1a2'] 0.77603894
['ka1b1', 'kb1a1'] 0.7748481
\end{verbatim}

Dissimilar:

\begin{verbatim}
['rf4h4', 'Rf4h4'] -0.6773491
['Rg2h2', 'rg2h2'] -0.6777805
['qe5g7', 'Qe5g7'] -0.6786175
['qb4b2', 'Qb4b2'] -0.68006134
['Qb5c5', 'qb5c5'] -0.6817629
['Qb5c4', 'qb5c4'] -0.6822547
['Rg4g5', 'rg4g5'] -0.68284607
['rh4f4', 'Rh4f4'] -0.6843391
['qc8e8', 'Qc8e8'] -0.6846573
['qb4d4', 'Qb4d4'] -0.6861543
['Rg5h5', 'rg5h5'] -0.6867346
['qc1e1', 'Qc1e1'] -0.6908391
['Rh5g5', 'rh5g5'] -0.6913012
['Qc5d5', 'qc5d5'] -0.6930292
['qg7f7', 'Qg7f7'] -0.6932398
['rh4g4', 'Rh4g4'] -0.69328135
['rh5h4', 'Rh5h4'] -0.69604784
['Qf8f7', 'qf8f7'] -0.699224
['qf7f8', 'Qf7f8'] -0.7042312
['Rg4h4', 'rg4h4'] -0.71785456
\end{verbatim}

What we see in the first list are moves made in the most similar games, meaning in those where the moves differ little from each other. These are likely to be "template" games, probably with a guaranteed draw or, at the very least, with known and consistently reproducible continuations.

In the most similar moves, the squares to which the piece moves after the move match. Trivial quasi-synonyms are not a bad thing; predictability of the result is a good confirmation that there is some comprehensible logic or natural system behind the result. This means that analyzing these results is indeed worthwhile.

It is also logical that dissimilar moves are those made by pieces of different colors.

I built a vector model on "lemmatized" moves, meaning those where the information about the square from which the piece moves is excluded. The source text consisted of move listings, so positions were not considered. The complete size of the model is only $720$ moves.

\section{Conclusion}

I don't see how this representation of chess data can be used productively. It's unlikely that these vector models will help engines or people choose the best move. But in a purely academic sense, it's clear that such methods of information representation capture something important about the very nature of the game, which doesn't necessarily lead to a win.

Observations have shown that linguistic data are richer and more diverse than chess data. Fewer than $10,000$ words in the model is far fewer than the tens of thousands of words in models built from linguistic data. The limitations of the board and the number of pieces restrict the possible configurations of the vector space, despite the seemingly overwhelming number of possible moves.

The semantics reflected in the vector model include the stage of the game (moves from the opening and from the endgame are different), differentiate piece colors, the squares from which and to which moves are made, and connect together moves made by the same piece (including capturing the fact that castling involves two pieces).

The model doesn't account for differences between various openings (the difference between \pawn\texttt{e2e4} and \pawn\texttt{e2e3} is minimal). 
All of this highlights not so much the productivity of the word2vec algorithm as the significance of the context that the algorithm relies on. Even in chess, with its importance of position, it turns out that just from the sequence of moves without considering the position, a considerable amount of information about the course of the game and the semantics of individual moves can be reconstructed.

\section*{Acknowledgements}
I extend my gratitude to the International Master in chess and PhD in linguistics Atle Gr{\o}nn, with whom I discussed early versions of this work, and who offered valuable ideas for understanding the resulting outcomes. At the same time, all mistakes and inaccuracies in this paper remain solely my responsibility.

\bibliographystyle{unsrtnat}
\bibliography{references} 

\end{document}